\documentclass[10pt,twocolumn,letterpaper]{article}

\usepackage{cvpr}              

\usepackage{graphicx}
\usepackage{amsmath}
\usepackage{amssymb}
\usepackage{booktabs}
\usepackage{times}
\usepackage{microtype}
\usepackage{epsfig}
\usepackage[table,xcdraw]{xcolor}
\usepackage{caption}
\usepackage{xcolor}
\usepackage{float}
\usepackage{placeins}
\usepackage{color, colortbl}
\usepackage{stfloats}
\usepackage{enumitem}
\usepackage{tabularx}
\usepackage{xstring}
\usepackage{multirow}
\usepackage{xspace}
\usepackage{url}
\usepackage{subcaption}
\usepackage{arydshln}
\usepackage{xcolor}
\usepackage{bbm}
\usepackage[hang,flushmargin]{footmisc}
\usepackage{pifont}
\usepackage[accsupp]{axessibility}

\usepackage{makecell}

\definecolor{cyan}{RGB}{102,184,178}
\definecolor{lightblue}{RGB}{142,181,224}

\newcommand{\et}[2]{${#1}^{\pm{#2}}$}

\newcommand{\etr}[2]{$\textcolor{cyan}{{\textbf{#1}}}^{\pm{#2}}$}
\newcommand{\etbb}[2]{$\textcolor{lightblue}{\textbf{{#1}}}^{\pm{#2}}$}

\definecolor{cvprblue}{rgb}{0.21,0.49,0.74}
\usepackage[pagebackref,breaklinks,colorlinks,citecolor=cvprblue]{hyperref}

\usepackage[capitalize]{cleveref}
\crefname{section}{Sec.}{Secs.}
\Crefname{section}{Section}{Sections}
\Crefname{table}{Table}{Tables}
\crefname{table}{Tab.}{Tabs.}

\def\paperID{4211} 
\def\confName{CVPR}
\def\confYear{2024}

\begin{document}

\title{OmniMotionGPT: Animal Motion Generation with Limited Data}

\author{
Zhangsihao Yang$^{1}$\thanks{Work done while interning at OPPO},
Mingyuan Zhou$^{2}$, 
Mnegyi Shan$^3$,
Bingbing Wen$^3$, 
Ziwei Xuan$^2$
\\ 
Mitch Hill$^2$, 
Junjie Bai$^2$, 
Guo-Jun Qi$^{2,4}$,
Yalin Wang$^1$
\\
$^1$Arizona State University, USA
\qquad
$^2$OPPO Seattle Research Center, USA
\\
$^3$University of Washington, USA
\qquad
$^4$Westlake University, China
\\
}
\maketitle

\begin{abstract}
Our paper aims to generate diverse and realistic animal motion sequences from textual descriptions, without a large-scale animal text-motion dataset. While the task of text-driven human motion synthesis is already extensively studied and benchmarked, it remains challenging to transfer this success to other skeleton structures with limited data. In this work, we design a model architecture that imitates Generative Pretraining Transformer (GPT), utilizing prior knowledge learned from human data to the animal domain. We jointly train motion autoencoders for both animal and human motions and at the same time optimize through the similarity scores among human motion encoding, animal motion encoding, and text CLIP embedding. Presenting the first solution to this problem, we are able to generate animal motions with high diversity and fidelity, quantitatively and qualitatively outperforming the results of training human motion generation baselines on animal data. Additionally, we introduce AnimalML3D, the first text-animal motion dataset with 1240 animation sequences spanning 36 different animal identities. We hope this dataset would mediate the data scarcity problem in text-driven animal motion generation, providing a new playground for the research community.

\end{abstract}

\section{Introduction}
\label{sec:intro}

\begin{figure}[tp]
\centering
\includegraphics[width=0.47\textwidth]{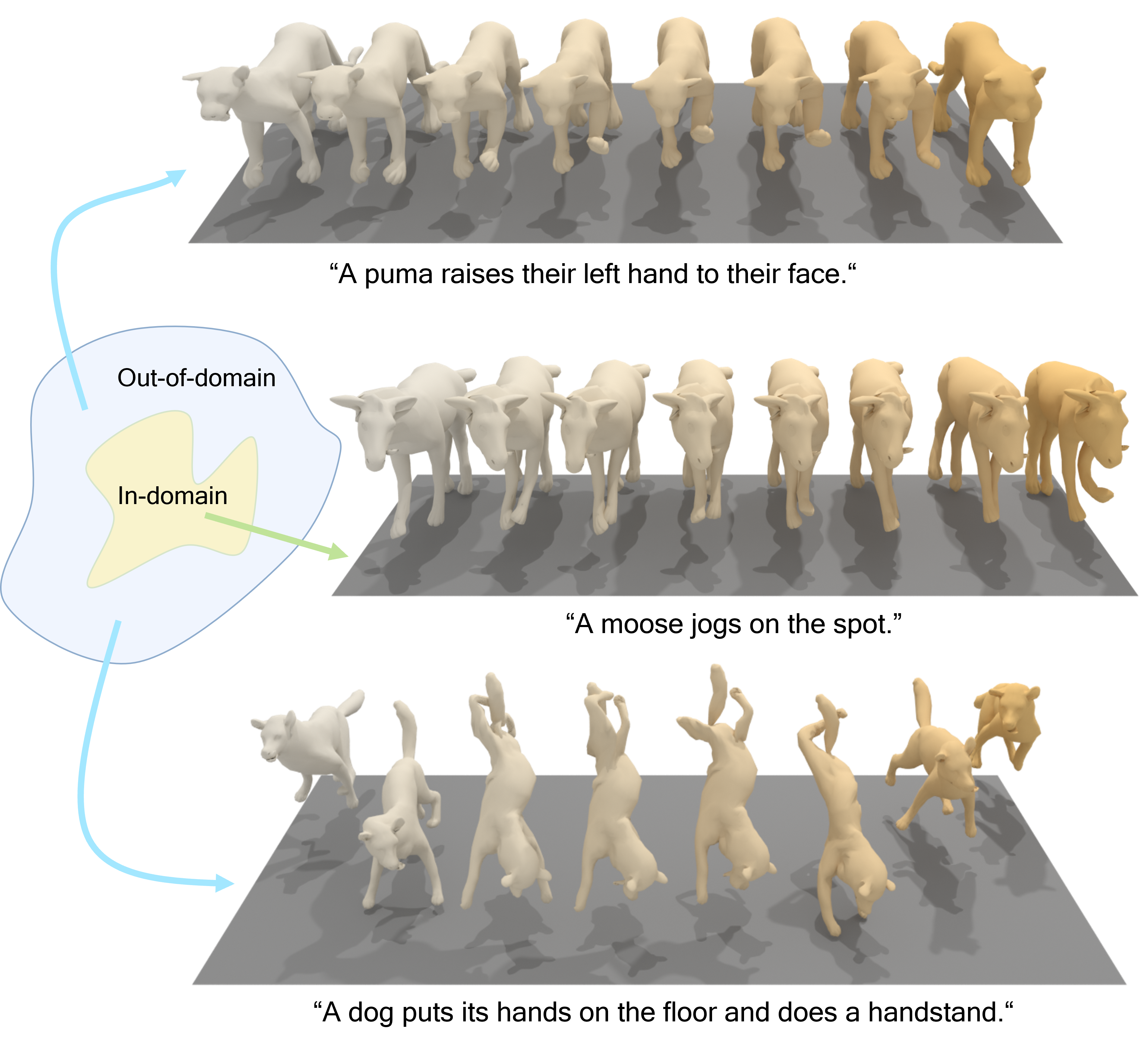}
\caption{
\textbf{Visualization of in-domain and out-of-domain motion generation from textual descriptions.}
Our model generates animal motion ranging from conventional movements to complex, out-of-domain behaviors.
The in-domain motion semantic latent space, highlighted by the yellow region, encapsulates common animal movements described in textual data. 
The out-of-domain latent space, delineated by the blue region, includes complex motions that are less frequently associated with animal behaviors, such as performing a handstand. The blue and green arrows denote our motion generation process from out-of-domain and in-domain prompts.
}
\label{fig:teaser}
\end{figure}

Computational modeling of 3D motions is an important topic with a wide range of applications, including robotics, virtual/mixed/augmented reality, gaming, and visual media.
Traditional methods for obtaining computational models of motions rely on human artists who use their observations of the real world to animate 3D assets \cite{li20214dcomplete}, or extensive motion capture process \cite{moeslund2001mocap}.
This process requires great effort and skill from artists or an expensive and time-consuming capture procedure.
Recent advances in generative modeling have led to breakthrough success for synthesizing realistic human motions using natural language textual descriptions \cite{tevet2023human, guo2022t2m, chuan2022tm2t, petrovich22temos, jiang2023motiongpt, tevet2022motionclip}.
Text-driven motion generation has the potential to greatly increase the efficiency and accessibility of motion animation. Despite the success of motion generation in the domain of human motions, significant obstacles remain which prevent similar techniques from being used to generate other kinds of motions.

In this work, we showcase a method to tackle the difficult problem of animal motion generation from text descriptions.
Text-driven animal motion generation is much less studied than human motion generation mainly due to dataset availability issues.
Animal motion data in the research community is very limited and not available at a comparable scale as human motion datasets \cite{plappert2016kit, guo2022t2m, lin2023motionx}. Specifically, there is no paired text-motion dataset for animal motion sequences at all, akin to HumanML3D \cite{guo2022t2m} in the human motion domain. This fundamental data scarcity problem motivates us to leverage information from human motions to supplement significantly smaller animal motion datasets.

To incorporate human motion data when training an animal motion model, we must address several key problems. Animals have different motion representations than humans, notably in terms of the number of joints and joint definitions \cite{zuffi20173d}. This makes it hard to directly transfer the knowledge from human motion models to animal ones. Moreover, human motion generators do not care too much about the skeleton information beyond joints \cite{guo2022t2m, tevet2023human}, while for animals, the skeleton offsets for different species could be different even if they share the same skeleton topology \cite{zuffi20173d}. Furthermore, animals perform much less diverse motion patterns than human beings in reality, even though animals are capable of mimicking most motion patterns of human beings. It is straightforward to collect a motion of hand clapping for human, but requires more effort for animals either in reality, which requires animal training, or in virtual, which requires the artists’ manual calibration of the animal arm movements.

To address the aforementioned challenges, we propose an architecture to transfer the knowledge from the human motion domain to enrich the generation of both in-distribution and out-of-distribution animal motions. 
We first design a transformer-based~\cite{vaswani2017attention} motion encoder that projects different skeletal motions to a primal joint's latent space which enables the translation between two different motion domains. 
By registering the motion both on a common textural space, we are able to connect human motion modality, language space, and animal motion modality, with CLIP \cite{radford2021learning} similarity loss.
We design three loss functions, latent consistency, CLIP similarity, and end-effector loss, to regularize the transformation of the latent feature from human motion to animal motion generation model. 
We additionally create the first animal language-motion dataset AnimalML3D for training and evaluation of our method. We generate skeleton motions and annotate textural descriptions for the existing DeformingThings4D \cite{li20214dcomplete} dataset that only contains animal motion mesh sequences. 

Our contribution can be summarized as follows:
\begin{itemize}[itemsep=0pt, topsep=3pt, leftmargin=10pt]
\item We present OmniMotionGPT,  a new framework that trains on sparse animal motion data and generates diverse motions from complex texts by transferring learned human motion knowledge.

\item We propose a new method to train motion autoencoders for both animal and human motion by aligning their semantic representation. Extensive experiments demonstrate that our method significantly outperforms existing methods both qualitatively and quantitatively.

\item We introduce AnimalML3D, the first dataset pairing text descriptions with 3D animal motions, which consists of 3720 human-written textual descriptions accompanying 1240 motions of 36 different animal identities. We hope our new dataset can provide a solid new playground for researchers interested in the animal text-motion task.
\end{itemize}

\begin{figure*}[t]
\includegraphics[width=\linewidth]{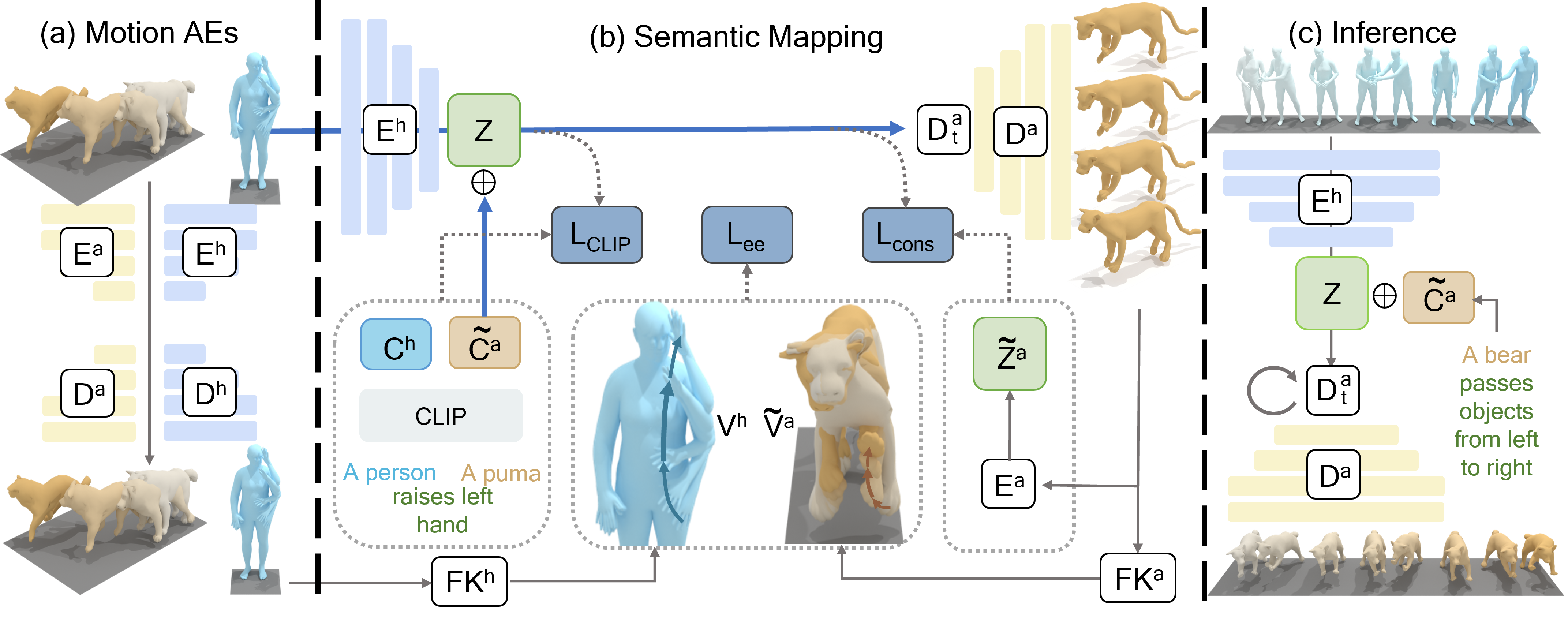}
\caption{
\textbf{The architecture of our training and inference stages.}
We train part (a) and part (b) at the same time. 
In (a), we train two motion autoencoders simultaneously, each within their domain, leveraging primal joints to maintain dimensional coherence in the latent space. 
Details on the structure and loss functions can be found in Section~\ref{sec:ae}. 
In (b), human motion is fed into the human motion encoder $E^h$ to produce a semantic-aware, subject-invariant latent code $\mathcal{Z}$. The CLIP feature of the subject-translated sentence and $\mathcal{Z}$ are concatenated together and passed into the animal text decoder $D_t^a$ and motion decoder $D^a$.
We introduce three losses to regularize the generated animal motions.
CLIP similarity loss $\mathcal{L}_{CLIP}$ extracts subject-invariant latent features. 
Latent consistency loss $\mathcal{L}_{cons}$ pushes the generated animal motion to be closer to the subject-invariant motion feature $\mathcal{Z}$.
End-effectors loss $\mathcal{L}_{ee}$ injects human motion velocity information into animals.
During inference in (c), we generate animal motions based on human motion sequences sampled from generative models.
Details on the architecture, loss functions, and inference process are elaborated in Section~\ref{sec:mapping}. 
}

\label{fig:pipeline}
\end{figure*}

\section{Related work}
\label{sec:related_work}

\paragraph{Animal Representations.} 
Several models have been developed to represent animal motion, including LASSIE~\cite{yao2022lassie}, SMAL~\cite{zuffi20173d}, and LASR~\cite{yang2021lasr,yang2021viser,yang2022banmo}.
SMAL and its enhanced variant SMALR~\cite{zuffi2018lions}, with more expressive features, extend of the widely-used human motion representation SMPL~\cite{loper2023smpl}, catering to the motion representations of five animal categories. 
LASR is introduced following SMAL to accommodate a broader range of animal species. 
LASSIE, along with its subsequent iteration Hi-LASSIE~\cite{yao2023hi}, employs a neural field around detected bones in images, but they are used more often in image or video reconstruction instead of motion generation.
Our approach utilizes SMAL as the core representation due to its explicit skeletal structure and the semantic meaning provided for each joint. 
Additionally, the compatibility of SMAL with the standard human motion representation SMPL \cite{loper2023smpl}, facilitates the knowledge transfer from human to animal motion distribution, which is crucial to our research.

\noindent\textbf{Human Motion Synthesis.} Human motion synthesis aims to generate diverse and natural 3D human motion. One major line of research focuses on motion generation based on existing motion frames. For example, predicting future motion from given frames \cite{mao2020history, fragkiadaki2015recurrent, bütepage2017deep, martinez2017motion, barsoum2017hpgan, hernandez2019human}, motion in-betweening \cite{duan2021singleshot, harvey2021recurrent, harvey2020inbetween, tang2022transition}, and motion generation from a simple sequence \cite{li2022ganimator}. Traditionally this has been modeled as a one-to-one relationship until recent generative models handle the stochastic nature of motion space and greatly increase the result diversity. Another line of work incorporates multimodal inputs as conditioning signals, including action label \cite{guo2020action2motion, petrovich21actor, wang2019learning}, music and audio \cite{huang2023dance, lee2019dance, tseng2023edge}, scene geometry \cite{Wang2021scene, wang2022humanise}, object interaction \cite{kulkarni2023nifty}, and text \cite{jiang2023motiongpt, tevet2022motionclip, zhang2022motiondiffuse, guo2022t2m, chuan2022tm2t}. Despite the amount of research effort in human motion generation, it remains an open problem whether such approaches could be migrated to other skeleton structures like animals, mainly due to the lack of datasets with comparable scales.

\noindent\textbf{Text-driven Human Motion Generation.} 
With the development of pre-trained language models, text-driven human motion synthesis becomes one of the most important conditional motion generation tasks. The goal is to synthesize realistic, diverse 3D human motion sequences that align semantically with given textual descriptions. MotionCLIP \cite{tevet2022motionclip} uses auto-encoder structures to learn a joint embedding of language and pose and thus generate animations. TEMOS \cite{petrovich22temos} and T2M \cite{guo2022t2m} leverage a VAE structure to map text into a normal distribution in the latent space. Later work TM2T \cite{chuan2022tm2t}, MotionGPT \cite{jiang2023motiongpt}, and T2MGPT \cite{zhang2023generating} learn to encode the motion sequences as discrete, quantized text/motion tokens in a fixed size codebook, and generate through an auto-regressive process. A parallel line of work utilizes diffusion model \cite{ho2020denoising} with text embedding as a condition. MDM \cite{tevet2023human} and MotionDiffuse\cite{zhang2022motiondiffuse} apply diffusion model to text-motion dataset through a transformer structure. ReMoDiffuse \cite{zhang2023remodiffuse} further integrates a retrieval mechanism to refine the denoising process. MLD \cite{chen2023executing} achieves better results and is two orders of magnitude faster than previous diffusion models by using the latent diffusion model. PhysDiff~\cite{yuan2023physdiff} further incorporates physical simulation to enforce realistic human motion rules. Nevertheless, the nature of diffusion models and VAEs requires a huge amount of data during training, and thus won't directly apply to animal motions. 

\noindent\textbf{Motion Retargeting.}
Many works in motion retargeting focus on transferring motion data between entities with topologically equivalent skeletons, particularly in human~\cite{gleicher1998retargetting,lee1999hierarchical,aberman2020skeleton} and animal contexts~\cite{maheshwari2023transfer4d}.
Some other works retarget human motion data to non-humanoid characters; these methods typically require humans to mimic animal motions~\cite{seol2013creature} or necessitate the creation of a paired dataset for motion transfer~\cite{abdul2017motion,yamane2010animating}.
Skeleton-free retargeting~\cite{wang2023hmc,liao2022skeleton,kulkarni2020articulation} is another emerging approach to retargeting 3D objects.
Our task differs from traditional retargeting as we directly generate motions from text descriptions.

\noindent\textbf{Motion and Pose Datasets.}
HumanML3D \cite{guo2022t2m} is built upon HumanAct12\cite{guo2020action2motion} and AMASS\cite{mahmood2019amass}, containing a broad range of human actions such as daily activities. Similarly, KIT language-motion dataset \cite{plappert2016kit} contains 3911 motions and 6278 natural language annotations. Motion-X \cite{lin2023motionx} is another large-scale 3D expressive whole-body motion dataset paired with textual annotations. On the other hand, for animals, we have Animal3D \cite{xu2023animal3d} which estimates static poses from animal images but doesn't contain dynamic motion sequences. DeformingThings4D \cite{li20214dcomplete} is perhaps the only animal motion dataset, but it's built for depth and optical flow estimation and therefore doesn't come with textual annotations and has a limited amount of motion sequences. To the best of our knowledge, there are no public animal text-motion datasets before us.

\section{Method}
\label{sec:method}

Our goal is to generate high-quality animal motions that are consistent with text descriptions. The overall training framework consists of two parts optimized simultaneously: motion autoencoder training for animals and humans, and joint training for knowledge transfer, as illustrated in Figure~\ref{fig:pipeline}. Section \ref{sec:ae} explains the separate training procedure of human motion and animal motion autoencoders. Section \ref{sec:mapping} describes the joint training mechanism that aligns human and animal motion spaces, along with integrating the text semantic latent space.  It also illustrates how this mechanism decodes human motion embedding to generate animal motion in the inference stage.

\subsection{Integrating Joint and Text Awareness in Motion Autoencoders}
\label{sec:ae}
\noindent\textbf{Motion Representation.} In object motion representation, the kinematics can be abstracted through a skeletal model. 
This skeletal structure is conceptualized as a tree graph, with joints as nodes and armatures as edges as defined in~\cite{aberman2020skeleton}.
The number of joints $J$ is consistently one greater than the number of armatures $A$.
We represent skeletal motion using a static component $\mathcal{S} \in \mathbb{R}^{(J-1) \times S}$, with $S$ as static features' dimensionality, usually set as a 3D vector ($S = 3$).
Beyond this static representation, our dynamic component comprises three parts: global rotation $\mathcal{R} \in \mathbb{R}^{T \times Q}$, global translation $\mathcal{T} \in \mathbb{R}^{T \times 3}$, and joint rotations $\mathcal{Q}\in \mathbb{R}^{T \times (J-1) \times Q}$ relative to their parents, excluding the root joint. 
We select \( Q=6 \), following \cite{zhou2019continuity}, to represent the rotations of each joint and global root.
After augmenting the global translation to a $Q$-dimensional vector by padding zeroes to it, the dynamic component can be represented as \( \mathcal{D} \in \mathbb{R}^{T \times (J +1) \times Q} \) by concatenating $\mathcal{R}$, $\mathcal{T}$, and $\mathcal{Q}$. 
$\mathcal{D}$ is a sequence of poses $\mathcal{P}_t \in \mathbb{R}^{(J +1) \times Q}$ at frame $t$.
Primal joints are the joints that have a degree not equal to 2 in the skeletal graph. 
Intersecting primal joints is the intersection of primal joints between skeleton graphs.

\begin{figure}[t]
\centering
\includegraphics[width=0.47\textwidth]{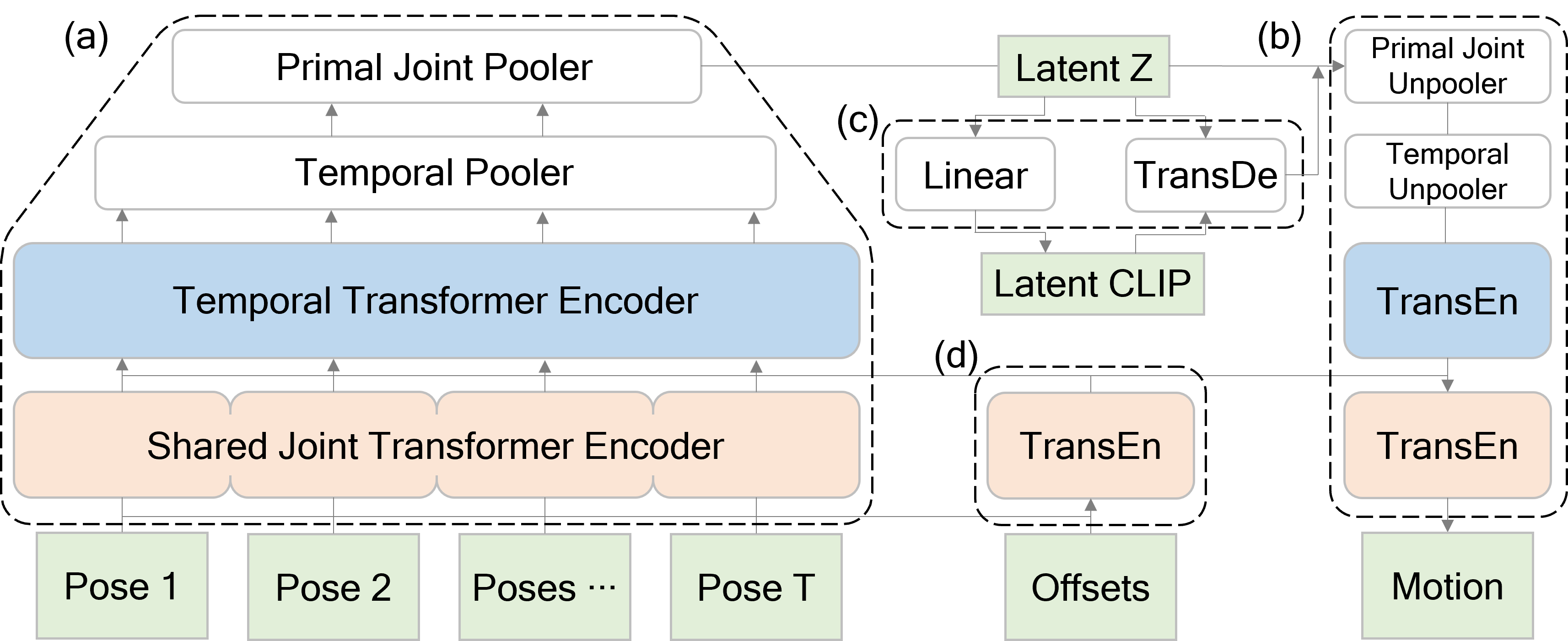}
\caption{
\textbf{Overview of the proposed motion autoencoder.} 
(a) shows the initial processing with the Temporal Transformer and Shared Joint Transformer Encoders. 
(b) illustrates the Primal Joint Unpooling and Temporal Unpooling sections. 
(c) represents the MLP and TransDe components leading to the Latent Z space. 
(d) indicates the final processing involving Latent CLIP and generation of motion and offsets.
}
\label{fig:autoencoder}
\end{figure}

\noindent\textbf{Joint-aware Motion Autoencoder.} Figure \ref{fig:autoencoder} shows an overview of our autoencoder model.
Our model begins with a transformer encoder extracting joint-level features from each pose.
The input is the concatenation of poses $\mathcal{P}_t \in \mathbb{R}^{(J+1) \times Q}$ and the corresponding, zero-padded static offsets $\mathcal{S'}\in \mathbb{R}^{(J+1) \times S}$. 
The shared joint transformer encoder generates a feature $\mathcal{F}_{j} \in \mathbb{R}^ {(J+1) \times f_j}$ for each pose. 
Similarly, another joint level transformer encoder is used to extract feature $\mathcal{F}_o \in \mathbb{R}^{(J+1) \times f_j}$ from $\mathcal{S}$.
Subsequently, a second transformer encoder extracts temporal features $\mathcal{F}_t \in \mathbb{R}^{T \times F_t}$, where $F_t = (J+1) \times f_t $, with concatenated input of $\mathcal{F}_j$ and $\mathcal{F}_o$.
Following this, a 1D pooling layer reduces the temporal dimension.
And a primal joint pooling layer selectively extracts features from intersecting primal joints (uniformly across different skeleton graphs) to form the latent feature $\mathcal{Z}=E(\mathcal{D}, \mathcal{S}) \in \mathbb{R} ^ {(T/l) \times J{p} \times f_z}$, where $l$ represents the temporal downsampling rate and $J_p$ is the number of primal joints. 
This is followed by a temporal unpooling layer, which replicates $\mathcal{Z}$ by a factor of $l$, and a joint unpooling layer that introduces zero-padding at non-primal joint locations.
Further refinement is executed via two transformer encoders, operating on temporal and joint dimensions similar to the initial encoding phase. 
The output, $\mathcal{F}_o = D(\mathcal{Z}, \mathcal{S})$, is formatted to match the dimensionality of the input dynamic $\mathcal{D}$.

\noindent\textbf{Text-aware Motion Autoencoder.}
To incorporate textual information into our autoencoder architecture, we develop a cross-modal encoding and decoding scheme. 
This involves encoding the latent vector $\mathcal{Z}$ into the CLIP feature domain $\mathcal{Z}_{CLIP} = E_t(\mathcal{Z})$, where $E_t$ is a latent encoder. 
Then we have a latent decoder $D_t$ to decode back to joint-aware latent space $\mathcal{Z}_t = D_t (\mathcal{C}, \mathcal{Z})$.
The decoder, a causal attention \cite{radford2018improving} based transformer, accepts both CLIP features and the latent vector $\mathcal{Z}$ as inputs. 
Its output, subsequently channeled into the joint-aware decoder to get $\mathcal{F}_{text}$, enables synchronous training of both autoencoder networks. 
This dual functionality facilitates the conversion of motion into CLIP representations and vice versa, extending the capabilities of the autoencoder to include sequential motion decoding from textual descriptions.

\noindent\textbf{Training Objectives.} There are three losses used to train the motion autoencoder: the reconstruction loss from the joint-aware autoencoder $\mathcal{L}_{jrec}=||P - \mathcal{F}_o||_2$; the CLIP similarity loss $\mathcal{L}_{CLIP} = 1 - \cos (\mathcal{Z}_{CLIP}, \hat{\mathcal{Z}}_{CLIP})$; and the CLIP forward reconstruction loss $\mathcal{L}_{trec} = ||P - F_{text}||_2$. 
The total loss to train motion autoencoder is 
\begin{equation}
\mathcal{L}_{ae} = \mathcal{L}_{jrec} + 
+ \lambda_1 \mathcal{L}_{CLIP} + 
\lambda_2 \mathcal{L}_{trec}
\label{formula:2}
\end{equation}
where $\lambda_1 = 1.0$ and $\lambda_2 = 1.0$ in our experiment.

\subsection{Semantic Mappings between Motion Autoencoders}
\label{sec:mapping}

\noindent\textbf{Architecture.} 
Our objective is to generate new animal motions by leveraging human motion data, which encompasses a wide range of types and semantic interpretations. 
We train two autoencoders: a human-focused model on abundant human motion data and an animal-focused model on the animal’s limited dataset enriched with latent features extracted from the human motion model. 

In Figure \ref{fig:pipeline}, static $\mathcal{S}^{h}$ and dynamic $\mathcal{D}^{h}$ components of human motions are encoded into a latent motion feature space $\mathcal{Z}^{h}$ through the human encoder $E^{h}$.  For simplicity, we use $h$ to represent $human$ and $a$ to represent $animal$.
We then replace the subject of the sentence describing the human motion with the name of the targeted animal. The CLIP embedding of the original sentence is $\mathcal{C}^h$ and the edited sentence is $\tilde{\mathcal{C}}^a$.
These features are subsequently passed into the animal motion decoders, $D^{a}$ and $D_{t}^{a}$, which incorporate the static components $\mathcal{S}^{a}$ of animal motions to generate the synthetic output $\tilde{\mathcal{F}}_o = D^{a}(D_t^{a}(\tilde{\mathcal{C}}^a, \mathcal{Z}^{h}), \mathcal{S}^{a})$.
we simplify this process as $\tilde{\mathcal{F}}_o = \tilde{D}^{a}(\tilde{\mathcal{C}}^a, \mathcal{Z}^{h}, \mathcal{S}^{a})$.

\paragraph{Training Objectives.} To supervise training of the aforementioned architecture, we design three loss functions, as illustrated in Figure \ref{fig:pipeline}.

\underline{CLIP Similarity Loss.} Our objective is to extract a subject-invariant latent feature $\mathcal{Z}^{h}$ from human motion data, encapsulating the action independent of the subject. 
For instance, the extracted latent feature $\mathcal{Z}^{h}$ of `a person is running' should encapsulate the notion of `running' exclusively, abstracting away from `a person'. 
We integrate this subject-invariant feature into our network by employing two distinct CLIP cosine similarity losses. 
The first loss function minimizes the distance between the CLIP feature $\mathcal{C}^h$ of the human motion sentence and $\mathcal{Z}^{h}$, as introduced in Section~\ref{sec:ae}. 
The second loss function minimizes the distance between the modified CLIP feature $\tilde{\mathcal{C}}^a$, obtained by substituting the subject in the sentence with an animal name, and $\mathcal{Z}^{h}$, represented as 
\begin{equation}
\mathcal{L}_{CLIP} = 1 - \cos (E_t^a({\mathcal{Z}}^{h}), \tilde{\mathcal{C}^a}).
\end{equation}
This dual loss strategy promotes subject-invariance in the latent feature $\mathcal{Z}^{h}$.

\underline{Latent Consistency Loss.} To ensure the integrity of the latent feature transformation within our framework, we define the Latent Consistency Loss, $\mathcal{L}_{cons}$.
This loss quantifies the discrepancy between the human latent feature ${Z}_{h}$ and its reconstructed counterpart obtained after processing through the animal motion decoder and encoder, $E^{a}(\tilde{D}^{a}(\tilde{\mathcal{C}}^a, \mathcal{Z}^{h}, \mathcal{S}^{a}), \mathcal{S}^{a})$. 
It is expressed as the L2 norm of their difference:  
\begin{equation}
\mathcal{L}_{cons} = ||{Z}^{h} - E^{a}(\tilde{D}^{a}(\tilde{\mathcal{C}}^a, \mathcal{Z}^{h}, \mathcal{S}^{a}), \mathcal{S}^{a})||_{2}.
\end{equation}


\underline{End-Effectors Loss.} Our End-Effectors Loss ensures that the dynamic translation of motion from humans to animals maintains kinematic integrity by comparing the velocities at the skeletal structure's extremities, known as end-effectors. 
These points, defined as terminal nodes on the skeleton graph, are crucial for generating realistic motion. 
Velocities for these points are computed using forward kinematics, $FK_{ee}$ (see Appendix for methodology). The velocity for human motion end-effectors is calculated as $\mathcal{V}^{h} = FK_{ee}(\mathcal{D}^{h}, \mathcal{S}^{h})$, and for synthetic animal motion as $\tilde{\mathcal{V}}^{a} = FK_{ee}(\tilde{\mathcal{F}}_o, \mathcal{S}^{a})$. 
The loss is defined by the L2 norm of the velocity difference: 
\begin{equation}
\mathcal{L}_{ee} = ||\mathcal{V}^{h} - \tilde{\mathcal{V}}^{a}||_2
\end{equation}
guiding the network to generate animal motions that reflect the dynamic properties of human movements.

The total loss function for cross-domain motion adaptation is represented as:
\begin{equation}
\mathcal{L}_{cross} = \lambda_3 \mathcal{L}_{cons} + \lambda_4 \mathcal{L}_{CLIP} + \lambda_5 \mathcal{L}_{ee}
\label{eq:cross}
\end{equation}
where $\lambda_3 = 0.1$, $\lambda_4 = 1.0$, and $\lambda_5=100$.
The training objective for the entire framework is thus represented by:
\begin{equation}
\mathcal{L}_{total} = \mathcal{L}_{ae}^{h} + \mathcal{L}_{ae}^{a} + \mathcal{L}_{cross}^{a}.
\label{eq:total}
\end{equation}

\paragraph{Inference.}
During the inference phase, our framework starts by converting a textual description into the corresponding CLIP feature $\tilde{\mathcal{C}}$. 
In parallel, a human motion—either from an existing motion generation method or from a ground truth motion—is encoded through $E^h$ to produce the latent human motion feature $\mathcal{Z}^h$. 
These features are inputs to the animal textual decoder $D^a_t$, which samples a new latent feature $\tilde{\mathcal{Z}}$. Then the feature is fed into the animal motion decoder $D^a$, generating the intended animal motion.

\section{AnimalML3D Dataset}
\label{sec:data}
To address the data scarcity problem, we introduce AnimalML3D, the first animal language-motion dataset which has 922 training pairs and 318 test pairs. 
It extends DeformingThings4D~\cite{li20214dcomplete} which consists of 1972 animation sequences spanning 31 different animals or humanoid categories with dense 4D annotation. 
We select motion sequences that correspond to the SMAL categories \cite{zuffi20173d}, and precisely extract skeletal data from the selected motions. 
This curation process resulted in a robust set of 1,240 animation sequences, which are then divided into a training set of 922 sequences (23 identities) and a test set of 318 sequences (13 identities). 

We introduce two significant enhancements to DeformaingThing4D.
First, we created three descriptive captions by a group of well-trained human annotators for each motion, generating a comprehensive dataset that consists of 3,720 sentences, with a minimum sentence length criterion of five words.
Second, we generated skeletal motion data derived from the original animations.

We first fit a SMAL template to the first frame of the mesh, employing the approach detailed in~\cite{biggs2019creatures}. 
While this initial step establishes an approximate starting alignment, it necessitates further refinement for a precise fit to the target mesh.
To achieve a more precise overlay with the target mesh, we utilized Wrap4D, a commercial software specifically designed for processing 4D sequences.
We determined corresponding keypoints, ranging from 10 to 30, on the fitted SMAL template and the target mesh geometry.
Having established this keypoint correspondence in the inaugural frame, Wrap4D is then employed to systematically morph the SMAL template across the entire sequence, ensuring that the adapted mesh conformed to the keypoint definitions and maintained the topological consistency of the SMAL model throughout the frames.
Subsequently, the joint positions were computed using the joint regression matrix as outlined in~\cite{loper2023smpl}. 
Comprehensive details of dataset curation and visual illustrations of the mesh quantities and procedural results are included in the Appendix.

\begin{table*}[t]
    \centering
    \scalebox{0.97}{

    \begin{tabular}{l c c c c c c c}
    \toprule
    \multirow{2}{*}{Methods}  & \multicolumn{3}{c}{R-Precision $\uparrow$} & \multirow{2}{*}{FID-OOD} & \multirow{2}{*}{MM-Dist $\downarrow$} & \multirow{2}{*}{Diversity $\uparrow$} & \multirow{2}{*}{MModality $\uparrow$}\\

    \cline{2-4}
    ~ & Top-1 & Top-2 & Top-3 \\
    
    \midrule

        T2M-GPT~\cite{zhang2023generating} &
        \et{ 0.089}{ .007}          & 
        \et{ 0.153}{ .007}          & 
        \et{ 0.214}{ .007}          & 
        \etr{2.792}{ .033}          & 
        \et{ 0.775}{ .004}          & 
        \etbb{44.761}{2.693}        & 
        \et{22.958}{0.731}          \\
  
        MotionGPT~\cite{jiang2023motiongpt} &
        \et{ 0.148}{ .008}                  & 
        \et{ 0.226}{ .008}                  & 
        \et{ 0.285}{ .008}                  & 
        \etbb{ 2.211}{ .034}                & 
        \et{ 0.741}{ .004}                  & 
        \et{44.334}{2.733}                  & 
        \et{13.967}{1.098}                  \\

        MDM~\cite{tevet2023human}   & 
        \et{0.336}{.010}            & 
        \et{0.523}{.012}            & 
        \et{0.649}{.014}            & 
        \et{1.167}{.027}          & 
        \et{0.501}{.003}            & 
        \etr{52.137}{2.690}         & 
        \et{22.108}{2.338}          \\

        MotionDiffuse~\cite{zhang2022motiondiffuse} & 
        \etbb{0.407}{.017}                          & 
        \etbb{0.614}{.015}                          & 
        \etbb{0.733}{.015}                          & 
        \et{1.019}{.014}                            & 
        \etbb{0.464}{.004}                          & 
        \et{38.821}{1.790}                          & 
        \etbb{31.350}{0.646}                        \\

    \midrule

        OMGPT (Ours)                & 
        \etr{0.850}{.009}           & 
        \etr{0.935}{.007}           & 
        \etr{0.964}{.006}           & 
        \et{1.453}{.021}            & 
        \etr{0.355}{.003}           & 
        \et{43.804}{1.701}          & 
        \etr{34.492}{0.874}         \\

    \bottomrule
    \end{tabular}
    }
    \vspace{-1mm}
    \caption{
    \textbf{Comparison with the state-of-the-art methods on out-of-distribution text descriptions.}
    We evaluate all methods using metrics from~\cite{guo2022t2m}.
    \texttt{FID-OOD} is used to gauge out-of-distribution performance, differentiating it from typical in-distribution assessments.
    We report each metric's average and standard deviation, based on 20 evaluations.
    The best and second-best results are highlighted in \textcolor{cyan}{cyan} and \textcolor{lightblue}{blue}.
    }
    \label{tab:out_dist}

\end{table*}

\begin{table*}[t]
    \centering
    \scalebox{0.97}{

    \begin{tabular}{l c c c c c c c}
    \toprule
    \multirow{2}{*}{Methods}  & \multicolumn{3}{c}{R-Precision $\uparrow$} & \multirow{2}{*}{FID $\downarrow$} & \multirow{2}{*}{MM-Dist $\downarrow$} & \multirow{2}{*}{Diversity $\uparrow$} & \multirow{2}{*}{MModality $\uparrow$}\\

    \cline{2-4}
    ~ & Top-1 & Top-2 & Top-3 \\
    
    \midrule

        \textbf{Real motion}    &
        \et{ 0.558}{ .049}      & 
        \et{ 0.734}{ .040}      & 
        \et{ 0.839}{ .032}      & 
        \et{ 0.105}{ .005}      & 
        \et{ 0.357}{ .006}      & 
        \et{22.795}{1.843}      & 
        -                       \\

    \midrule
        T2M-GPT~\cite{zhang2023generating} &
        \et{ 0.080}{ .024}          & 
        \et{ 0.168}{ .023}          & 
        \et{ 0.248}{ .042}          & 
        \et{ 1.084}{ .042}          & 
        \et{ 0.636}{ .013}          & 
        \etbb{33.403}{1.902}        & 
        \etr{20.078}{1.096}         \\

        MotionGPT~\cite{jiang2023motiongpt} &
        \et{ 0.142}{ .016}                  & 
        \et{ 0.233}{ .032}                  & 
        \et{ 0.307}{ .042}                  & 
        \et{ 0.748}{ .050}                  & 
        \et{ 0.558}{ .010}                  & 
        \et{29.265}{2.453}                  & 
        \et{10.311}{1.537}                  \\

        MDM~\cite{tevet2023human}   & 
        \et{ 0.379}{ .051}          & 
        \et{ 0.554}{ .058}          & 
        \et{ 0.646}{ .048}          & 
        \et{ 0.505}{ .038}          & 
        \et{ 0.487}{ .008}          & 
        \et{27.826}{1.643}          & 
        \et{13.593}{1.038}          \\

        MotionDiffuse~\cite{zhang2022motiondiffuse} & 
        \etbb{ 0.505}{ .037}                        & 
        \etbb{ 0.695}{ .045}                        & 
        \etbb{ 0.805}{ .041}                        & 
        \etbb{ 0.401}{ .024}                        & 
        \etbb{ 0.421}{ .007}                        & 
        \et{25.194}{1.510}                          & 
        \et{ 7.081}{0.357}                          \\

    \midrule
        OMGPT (Ours)        &
        \etr{ 0.539}{ .064} & 
        \etr{ 0.721}{ .063} & 
        \etr{ 0.830}{ .043} & 
        \etr{ 0.223}{ .036} & 
        \etr{ 0.348}{ .007} & 
        \etr{37.487}{1.575} & 
        \etbb{17.487}{0.792}\\

    \bottomrule
    \end{tabular}
    }
    \vspace{-1mm}
    \caption{
    \textbf{Comparison with the state-of-the-art methods on our AnimalML3D test set.} 
    Methods are evaluated using metrics from~\cite{guo2022t2m}, with top results in \textcolor{cyan}{cyan} (best) and \textcolor{lightblue}{blue} (second-best).
    We report each metric's average and standard deviation, based on 20 evaluations.
    }
    \label{tab:in_dist}

\end{table*}

\section{Experiments}
\label{sec:exp}

\noindent\textbf{Baselines and Evaluation Settings.} We compare our model performance with various motion generation models, including T2MGPT~\cite{zhang2023generating}, MotionGPT~\cite{jiang2023motiongpt}, MDM~\cite{tevet2023human} and MotionDiffuse~\cite{zhang2022motiondiffuse}. T2MGPT and MotionGPT employ a two-stage pipeline with VQVAE~\cite{oord2017vae} and GPT~\cite{brown2020gpt}, whereas MDM and MotionDiffuse utilize a single-stage diffusion model. All models are trained on the proposed AnimalML3D dataset.

We evaluate the results on two tasks. In in-distribution (ID) setting, we generate with prompts from the AnimalML3D dataset. In out-of-distribution (OOD) setting, we generate with prompts from the HumanML3D dataset by replacing the subject phrase with an animal name. 

We use the same set of evaluation metrics as in \cite{guo2022t2m}. 
\textit{R-precision} measures retrieval accuracy by comparing the input text to the generated motions. 
\textit{Frechet Inception Distance (FID)} measures the distance between generated motion distribution and testing motion distribution for ID experiments. As there is no ground truth animal motion for OOD experiments, we compare the distance between generated OOD motions and whole ground truth Animal3D dataset to compute the FID-OOD metric. 
\textit{Multimodal Distance (MM-Dist)} gauges the distance between the generated motion and the corresponding sentences in the latent space, using the outputs from the human latent encoder $E_t^h$ and CLIP features.
\textit{Diversity} evaluates the differences between independently sampled motions. 
\textit{Multimodality (MModality)} assesses the variance within multiple motions generated from a single text description.


\noindent\textbf{Implementation Details.} 
We use a two-layer transformer with a dimension of 16 for the joint encoder/decoder, and a two-layer transformer with a dimension of 256 for the temporal encoder/decoder. The latent encoder $E_t$ head is a linear layer with an input size of $49 \times 7 \times 16$ and the caption decoder is a four-layer transformer decoder with a dimension of 256. We train with the total loss described in Section \ref{sec:method} in an end to end manner for 30000 steps. We use an Adam optimizer with learning rate $lr = 10^{-4}$,  betas $\beta = (0.9, 0.999)$, batch size $B=256$, exponential moving constant $\lambda=0.99$. 

We configure the SMPL and SMAL representations with 22 and 35 joints respectively. For the HumanML3D dataset, following the data processing step in \cite{guo2022t2m}, we only keep motion sequences between 20 and 196 frames. For the AnimalML3D dataset, given its smaller size, we only keep motion sequences between 10 and 196 frames.
For details on the convergence of each loss component, readers are referred to the Appendix.


\begin{figure*}[tp]
\centering
\includegraphics[width=0.92\textwidth]{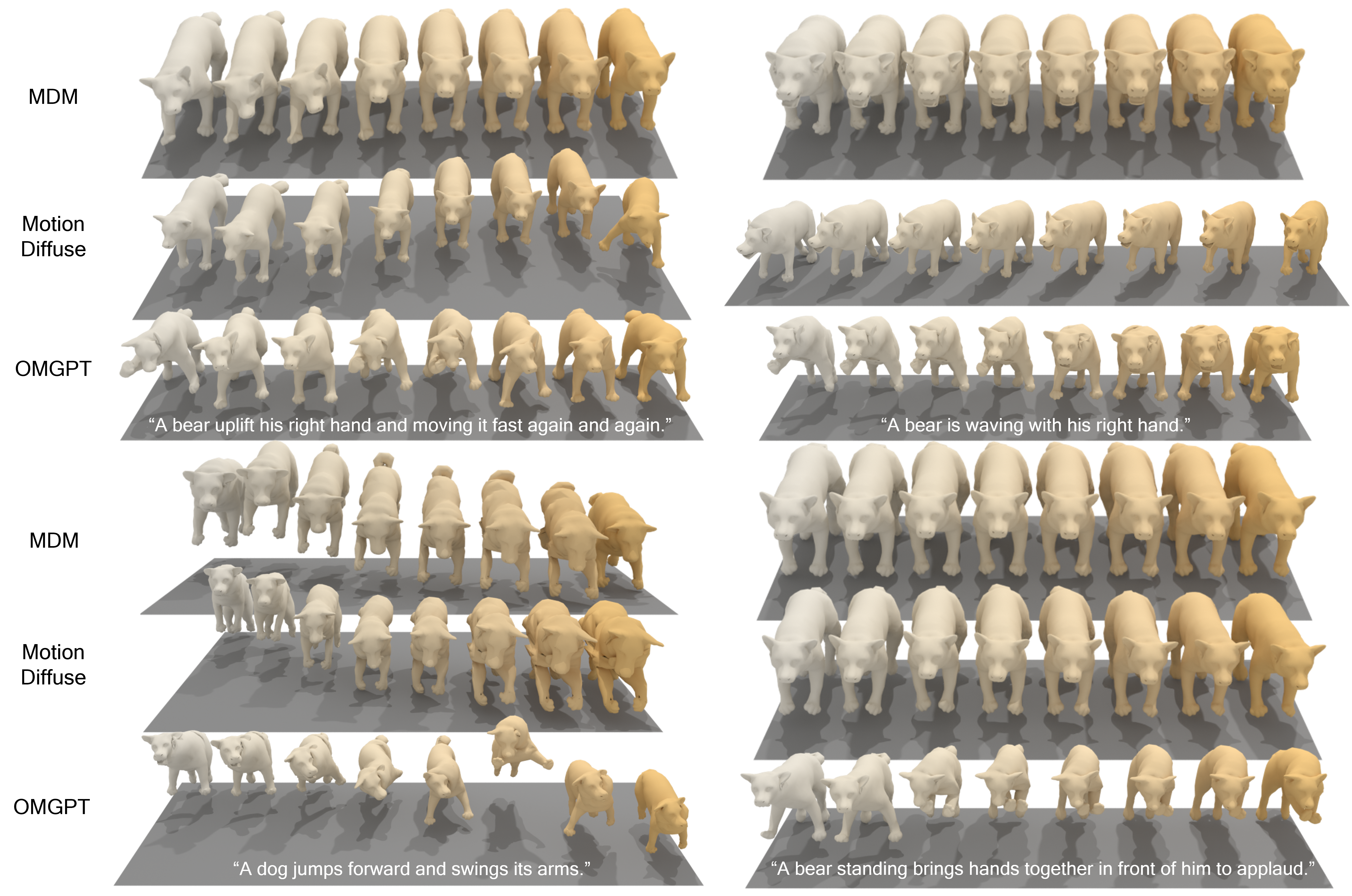}
\caption{
\textbf{Visual comparison between our method, OMGPT, and other baselines.}
Motions are generated according to the captions shown in the figure, evenly arranged in rows from left to right, showcasing a progression from beginning to end.
Our method demonstrates enhanced versatility and adherence to captions, outperforming baselines MDM~\cite{tevet2023human} and MotionDiffuse~\cite{zhang2022motiondiffuse}.
We assess performance using text descriptions adapted from HumanML3D~\cite{guo2022t2m}, modified to replace human subjects with various animals.
Notably, our method effectively processes OOD caption inputs, demonstrating significant improvements in alignment to these captions. Meanwhile, baselines are less adept at responding to such text descriptions.
}
\label{fig:baseline}
\end{figure*}

\noindent\textbf{Quantitative Motion Generation Comparison.}
Table \ref{tab:out_dist} and Table \ref{tab:in_dist} show our quantitative animal motion generation results, on both ID and OOD prompts. ID prompts are taken from our AnimalML3D test set, while out-of-distribution prompts are annotations from the HumanML3D test set with the subject replaced by an animal name. We compare our results with four recent human motion generation baselines. 
Two of them are based on VQVAE and GPT~\cite{zhang2023generating, jiang2023motiongpt} and the other two are based on diffusion models~\cite{zhang2022motiondiffuse,tevet2023human}.


The GPT-based models~\cite{zhang2023generating,jiang2023motiongpt} exhibit low R-precision scores for both ID and OOD experiments, attributed to the sparse training dataset of only 922 motion sequences and a relatively large codebook size of 512. 
This significant discrepancy leads to two key issues. First, the VQVA tends to overfit, reducing its ability to generalize in ID motion generation. 
Secondly, the large codebook size complicates the training of the transformer decoder, making it prone to generating repetitive motion patterns or noisy motions due to data sparsity (more details in the Appendix). 
This observation underscores that, although FID-OOD and Diversity scores are high, the extensive codebook size frequently results in unrealistic or repeated motions.
In contrast, our method, without relying on a fixed-size codebook, effectively handles small datasets with limited motion diversity.

Our OMGPT model outperforms the diffusion-based models MDM~\cite{tevet2023human} and MotionDiffuse~\cite{zhang2022motiondiffuse} in all metrics, both ID and OOD. While these models produce slightly higher R-precision and lower diversity scores compared to the GPT-based baselines, indicating better robustness to small datasets and text-motion alignment, they fall short in generating diverse motions from OOD prompts.

Additionally, note that OMGPT's superiority on OOD prompts is more prominent than ID prompts. This is because of its ability to incorporate human motion knowledge into the training process, and thus adaptable to a wider range of potential prompts. 
It is infeasible to jointly train with human data in all four baseline methods due to the motion representation difference in nature. 

\noindent\textbf{Qualitative Motion Generation Analysis.}
Figure \ref{fig:baseline} presents our generated motion sequences in comparison with baseline approaches MDM~\cite{tevet2023human}
and MotionDiffuse \cite{zhang2022motiondiffuse}.  With abundant knowledge transferred from human motion datasets, our model is able to generate results with better fidelity, alignment with the textual inputs, and diversity in complex motion descriptions.

Our OMGPT model outperforms baseline methods in three aspects. 
First, OMGPT demonstrates the ability to generate OOD motions that are out of the existing animal data distribution but in the human motion distribution.
The bottom right and top right examples show a bear clapping and waving hands which could be faithfully and reasonably generated by incorporating human motion knowledge with our framework but rarely happens in reality.
Second, OMGPT is able to comprehend a broader range of motion patterns not appearing in the animal dataset, like `fast' and `again and again' in the top left example. 
Third, OMGPT is capable of capturing complicated and composite motion descriptions, despite being built on an animal motion dataset with limited motion diversity and relatively simple prompts. 
The bottom left example illustrates OMGPT generating a sequence of motions (`jumping' and then `swinging arms') whereas the baseline methods are not able to handle.

\begin{table}[t]
\centering
\resizebox{0.48\textwidth}{!}{
\begin{tabular}{ccccc}
\toprule
\multirow{2}{*}{Exp}                                        & 
\multirow{2}{*}{\makecell{Configuration \\ Difference}}     &
\multirow{2}{*}{\makecell{R-Precision \\ Top-1 $\uparrow$}} &
\multirow{2}{*}{MM-Dist $\downarrow$}                       & 
\multirow{2}{*}{Diversity $\uparrow$}                       \\

\\

\midrule

A                   &
MLP Mapping         &
\et{0.351}{.009}    &
\et{0.476}{.002}    &
\et{31.406}{1.137}  \\

B                   &
$E_t$: MLP          &
\et{0.404}{.013}    &
\et{0.466}{.003}    &
\et{38.412}{1.900}  \\

C &
$\lambda_5 \mathcal{L}_{ee} = 0$    &
\et{0.477}{.017}                    &
\et{0.468}{.003}                    &
\et{\textbf{51.441}}{2.107}                  \\

D &
$\lambda_3 \mathcal{L}_{cons} = 0$  &
\et{0.508}{.019}                    &
\et{0.452}{.003}                    &
\et{43.946}{2.075}                  \\

\midrule

E                   &
-                   &
\et{\textbf{0.850}}{.009}    &
\et{\textbf{0.355}}{.003}    &
\et{43.804}{1.701}  \\

\bottomrule
\end{tabular}
}
\caption{
\textbf{Ablation Study on the configurations of our framework.}
In our ablation study, we evaluate various configurations of our framework in comparison to the fully integrated model, with a focus on architectural choices and loss weights. 
The impact of these elements is assessed using metrics including OOD R-Precision, MM-Dist, and Diversity.
Each metric is evaluated 20 times to compute the average and standard deviation.
These results demonstrate the essential role of each design component in our model in achieving optimal performance.
}
\label{tab:ablation}
\end{table}


\begin{figure}[t]
\centering
\includegraphics[width=0.47\textwidth]{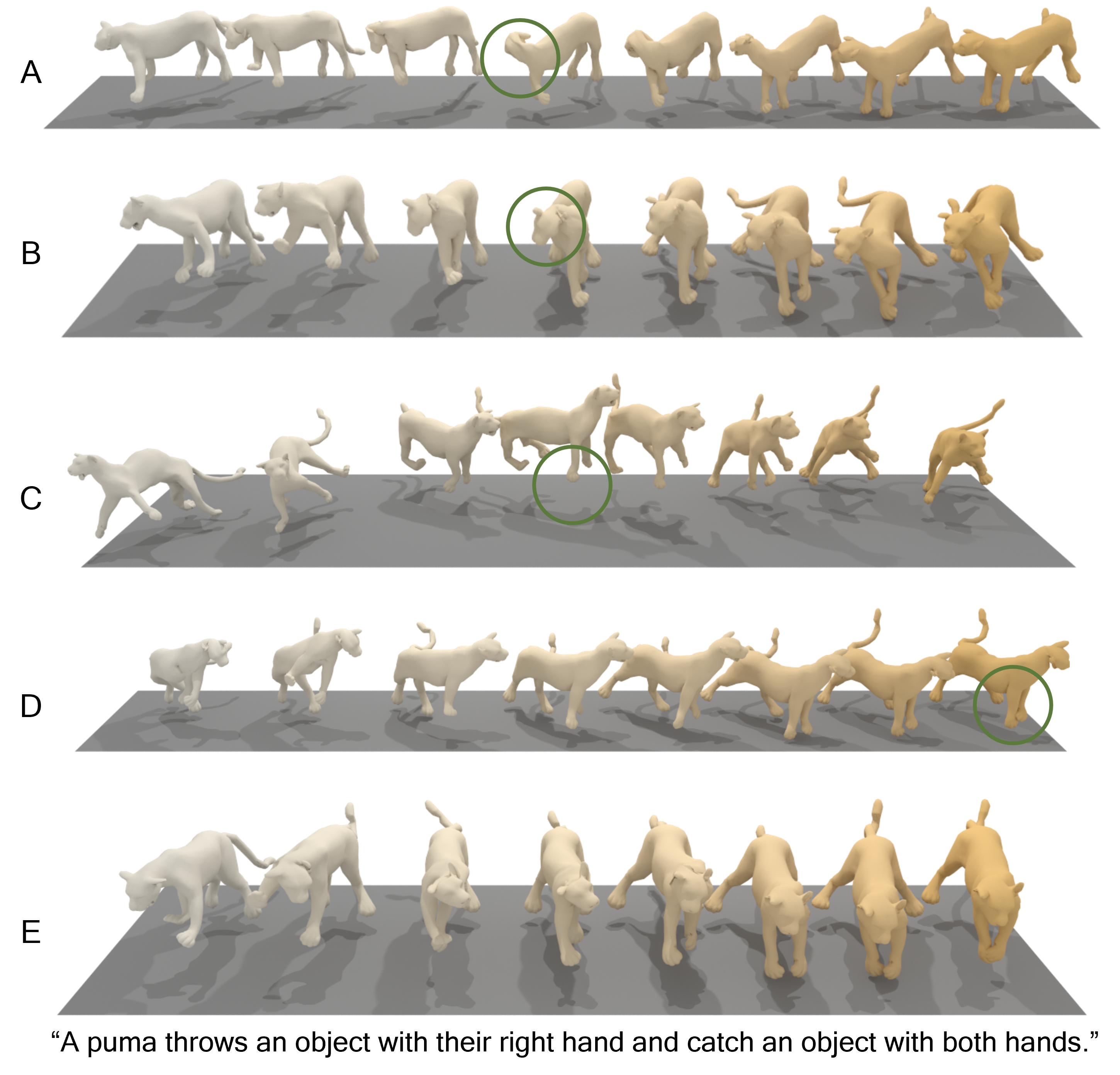}
\caption{
\textbf{Visualization of generated motion under different configurations.}
The letters \texttt{A-E} correspond to the \texttt{Exp} identities in Table~\ref{tab:ablation}.
Motions are generated according to the caption shown in the figure.
The green circles highlight the unrealistic parts in the motions by making changes to the configurations of the designed framework.
Motions are evenly arranged in rows from left to right, showcasing a temporal progression from beginning to end.
}
\label{fig:ablation}
\end{figure}

\noindent\textbf{Ablation Study.}
To validate the effectiveness of our designed semantic mapping configuration, we present ablation studies in Table~\ref{tab:ablation}. 
We alter the structure and loss weights of our final model to analyze their impact on motion generation quality and visual representation, as shown in Figure~\ref{fig:ablation}.

\underline{Architecture} (\texttt{Exp A \& B}). \texttt{Exp A} shows that adding an MLP mapping between the human latent space and the generated animal latent space results in less dynamic motion, as illustrated in row A of Figure \ref{fig:ablation}. 
\texttt{Exp B} shows that altering the semantic head from a linear layer to MLP allows more flexibility in the latent space. 
However, this indirectly affects the motion latent, leading to reduced movement in some joints, as observed in row B of Figure \ref{fig:ablation}.

\underline{Loss Weight} (\texttt{Exp C \& D}). \texttt{Exp C} sets the weight for $L_ee$ to 0 and achieves higher motion diversity but at the expense of realism. 
Without the end effector loss, the generated motion appears unnaturally elevated above the ground, as shown in row C of Figure~\ref{fig:ablation}. 
\texttt{Exp D} demonstrates that omitting consistency loss leeads to incomplete motion sequences. This is evident in the `catch an object' sequence, where the final part is missing in the generated motion, as depicted in row D of Figure~\ref{fig:ablation}.

\section{Conclusion}
\label{sec:con}

In this work, we propose the first text-driven animal motion generation algorithm. We design a one-stage jointly-training architecture that first trains motion autoencoder for both animal and human domains and simultaneously trains a knowledge mapping mechanism to generate animal motion with human motion encodings. We demonstrate diverse and realistic animal motion generation results and present metrics quantitatively surpassing all baseline methods. Moreover, we contribute the first animal text-motion dataset AnimalML3D, creating a new playground to encourage future investigation in the field of animal motion generation. 


{\small
\bibliographystyle{ieee_fullname}
\bibliography{main}
}

\clearpage
\appendix


\section{Configurations of Joints}

\begin{figure*}[t]
\centering
\includegraphics[width=0.92\linewidth]{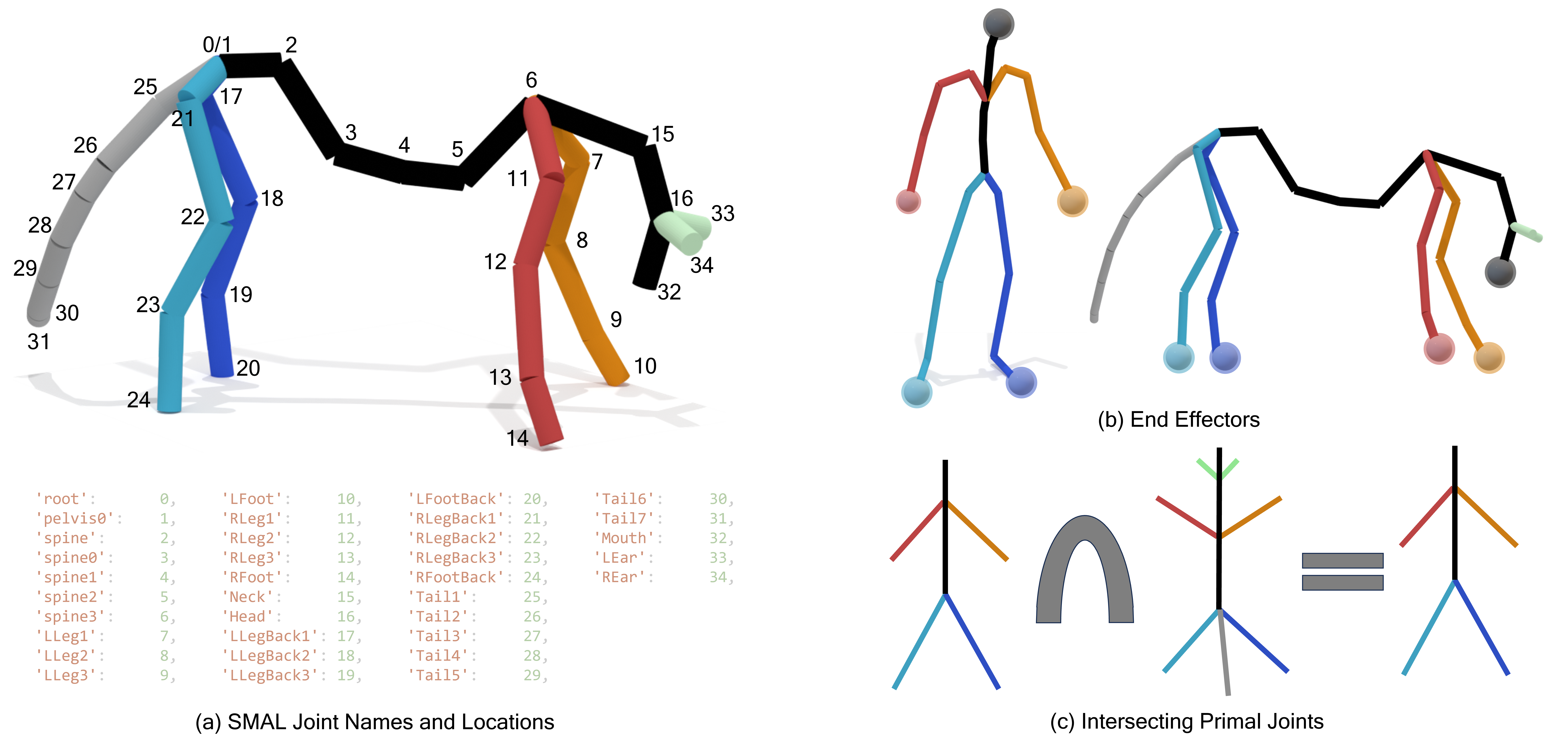}
\caption{
\textbf{Illustration of joints-related information.}
In part (a), we present the skeleton of the SMAL model~\cite{zuffi20173d}, including the names and indices of the joints. 
Part (b) displays the locations of the end-effectors in both the SMPL and SMAL models, represented by spheres of the same color for corresponding joints. In part (c), we depict the process of intersecting the primal skeleton graphs of SMPL and SMAL, illustrating the resulting intersecting primal skeleton between the two models.
}
\label{fig:supp_joints}
\end{figure*}

In part (a) of Figure~\ref{fig:supp_joints}, we outline the skeletal structure of the Skinned Multi-Animal Linear (SMAL) model ~\cite{zuffi20173d}.
The SMAL skeleton is comprised of 35 joints, notably with the ``root'' and ``pelvis0'' joints situated at the same location. 
A key distinction between the SMAL model and the Skinned Multi-Person Linear (SMPL) model~\cite{loper2023smpl} lies in the addition of a tail in SMAL, an element absent in the SMPL model.

We define essential concepts such as ``end effectors'', ``primal joints'', and ``intersecting primal joints'' in Section~\ref{sec:method}. 
These concepts are visually elaborated upon in Figure~\ref{fig:supp_joints}. 
For instance, in part (b) of the figure, we illustrate the end-effector joints for both SMAL and SMPL models, each marked with distinct color spheres to denote the five end-effector joints in both models.

Part (c) of Figure~\ref{fig:supp_joints} showcases the intersection of the primal skeletons of SMPL and SMAL. 
This intersection is subject to potential ambiguity. 
For example, the left leg branch in the SMPL graph could correspond to multiple components in SMAL, such as the left back leg, the tail, or even the right leg branch. 
Our approach aligns these intersections based on their semantic meanings, ensuring a meaningful and contextually appropriate mapping. 
The intersecting primal joints are clearly indicated in the figure, providing a nuanced understanding of the skeletal overlaps between the two models.

\section{Details of Data Processing}

\label{supp_sec:data_processing}

\begin{figure*}[t]
\centering
\includegraphics[width=0.92\linewidth]{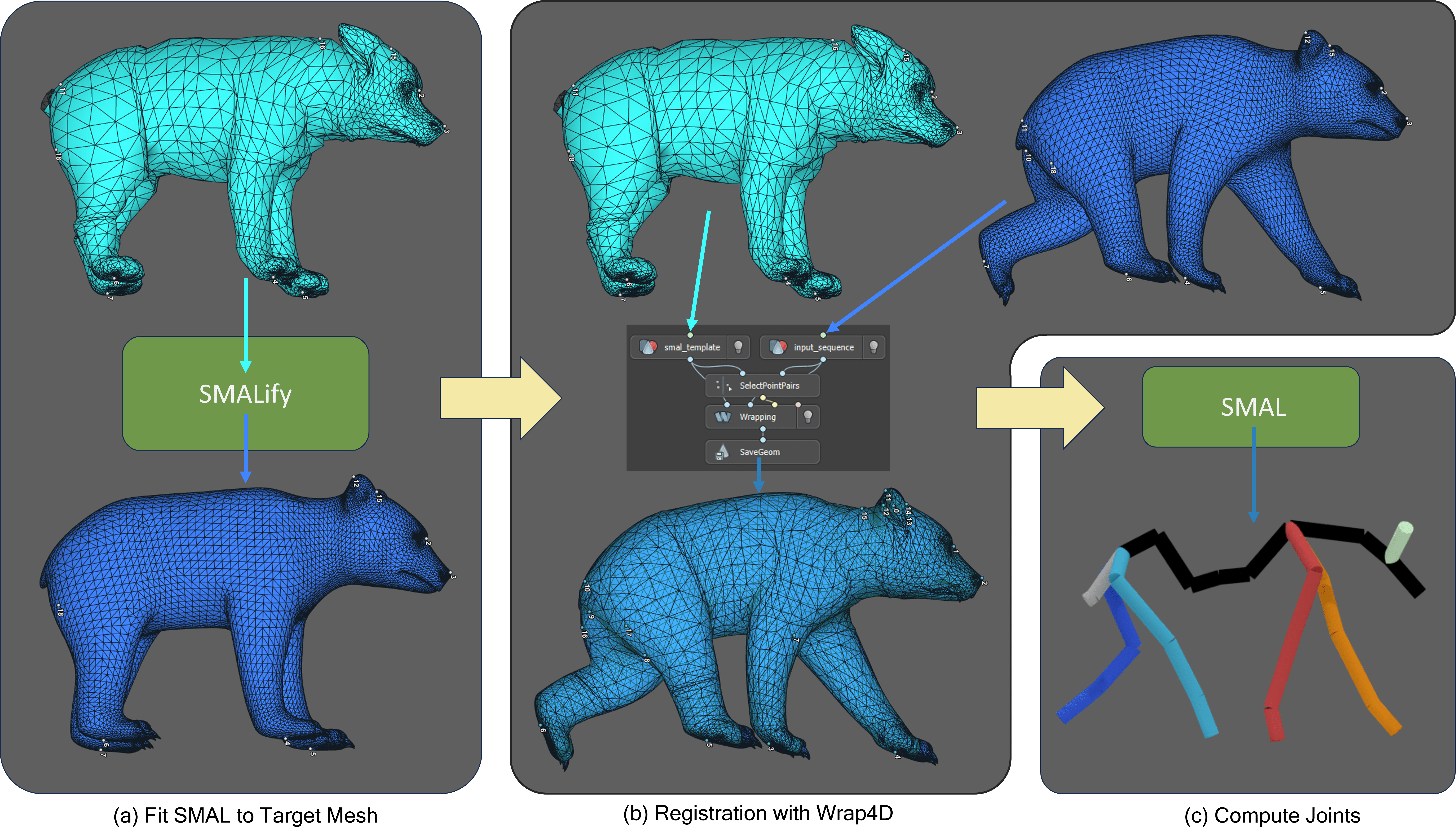}
\caption{
\textbf{Data processing pipeline for our AnimalML3D dataset.}
Our data processing pipeline is delineated into three stages: (a) fitting the SMAL model~\cite{zuffi20173d} to the target mesh, (b) registering the fitted mesh to a sequence of motions, and (c) computing joint positions from the registered mesh. 
In stage (a), we illustrate the target mesh (at the bottom) and the resulting fitted mesh (at the top). 
For stage (b), inputs to Wrap4D include the fitted mesh alongside the target mesh sequence (top right), with the output being the registered mesh maintaining SMAL topology (bottom), where white dots signify the corresponding points utilized for registration. 
In stage (c), we calculate the joint positions from the registered mesh; the figure highlights a short tail representation, typical of bear species where the tail is not prominently visible.
}
\label{fig:supp_data_processing}
\end{figure*}

In Figure~\ref{fig:supp_data_processing}, we illustrate the three-stage data processing workflow for our AnimalML3D dataset, using a representative example. 
The initial stage involves fitting a SMAL model~\cite{zuffi20173d} to the animal's identity in the first frame, typically in a resting pose as depicted in the lower section of (a) in Figure~\ref{fig:supp_data_processing}. 
Our approach is developed upon the framework established by~\cite{biggs2019creatures}, with a notable modification replacing the losses with Chamfer Distance~\cite{achlioptas2018learning}. 
We build upon the framework presented by~\cite{biggs2019creatures}, incorporating a significant adaptation: we employ the Chamfer Distance as our loss function, as described by~\cite{achlioptas2018learning}, instead of the original loss terms used in~\cite{biggs2019creatures}.
The model optimization targets four parameters: scale ($S$), global translation ($T$), and the SMAL model parameters $\beta$ and $\theta$, which are refined using the Chamfer Distance~\cite{achlioptas2018learning} between points sampled from the computed mesh of the SMAL model and the target mesh, with 3000 points sampled per iteration. 
Optimization is executed in two phases using the Adam optimizer~\cite{kingma2014adam} with a learning rate of 0.005: 
initially, $S$ and $T$ are optimized over 50 epochs, followed by a comprehensive optimization of $S$, $T$, $\beta$, and $\theta$ for an additional 400 epochs to obtain the final mesh.

In the second stage, we utilize the software Wrap4D for mesh registration, aligning the roughly fitted mesh from the previous stage to the meshes of each frame. 
The blueprint code for this process is depicted in part (b) of Figure~\ref{fig:supp_data_processing}. 
Within the software environment, we establish corresponding points between the fitted mesh and the target mesh. 
For every unique identity in the dataset, we generate a distinct correspondence map, culminating in a total of 36 correspondence mappings required to process the entire dataset.

In the third stage, which is elaborated upon in Section~\ref{sec:data} of the main paper, we apply the joint regression matrix to the vertices of the SMAL model that preserve the topology. 
This application yields the positional data for the joints.


\section{Loss Details and Convergence}

\begin{figure*}[t]
\centering
\includegraphics[width=0.92\linewidth]{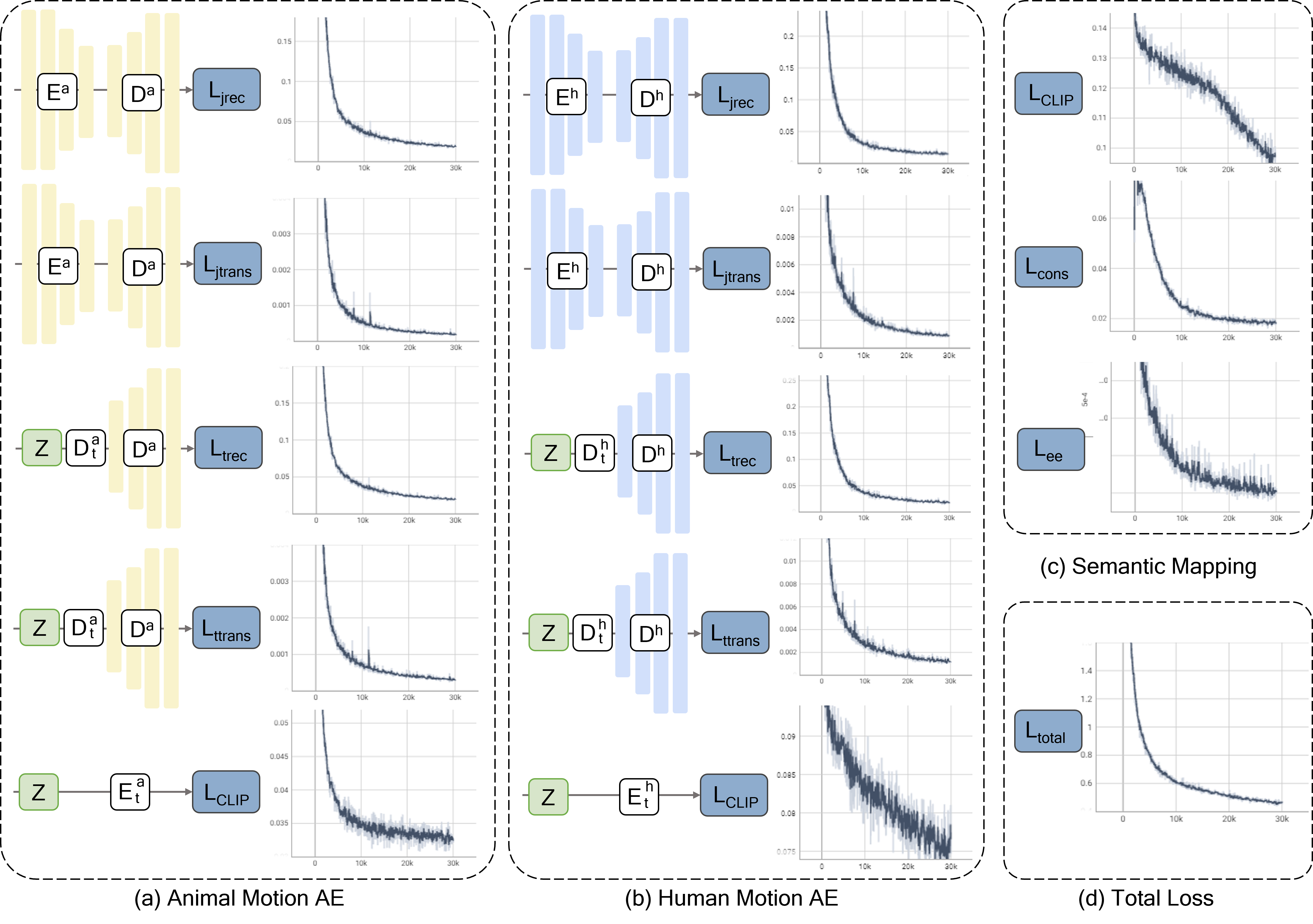}
\caption{
\textbf{Visualization of computation of loss functions and their convergence.}
Parts (a) and (b) illustrate the loss functions defined in Section~\ref{sec:ae}. 
Part (c) showcases the specific loss function introduced in Section~\ref{sec:mapping}. 
Finally, part (d) depicts the overall convergence of the total loss, represented as a weighted sum of all individual loss functions.
}
\label{fig:supp_loss}
\end{figure*}

In addition to the losses defined in Sections~\ref{sec:ae} and \ref{sec:mapping}, we introduce another loss function that employs global translation $\mathcal{T}$ to regularize generated motion. 
This loss is applied to both motions generated from the joint autoencoder and the text autoencoder, with a weight of 1.0. 
Empirically, we observed that incorporating global translation results in smoother motion generation, significantly reducing the shaking effect.

Figure~\ref{fig:supp_loss} illustrates the convergences of all the losses. 
Notably, the semantic loss $\mathcal{L}_{CLIP}$ does not converge close to 0.
There are two primary reasons for this.
First, achieving complete alignment between the motion and CLIP features is challenging. The motion encompasses attributes like velocity and facing direction, which are not fully captured in the CLIP features. 
Additionally, the CLIP features encode semantic nuances, such as differentiating between ``run'' for first and second-person pronouns and ``runs'' for third-person pronouns.
These disparities hinder a full alignment between motion and CLIP features. 
Second, our use of cosine similarity as a metric reveals that when similarity falls below 0.75, the resulting r-precision is approximately 63\%, a respectable rate in motion recall. 
This outcome underscores the nuanced relationship between motion and CLIP features, suggesting that perfect alignment may not be necessary for effective motion synthesis.


\section{More Our Results}

\begin{figure*}[t]
\centering
\includegraphics[width=0.97\linewidth]{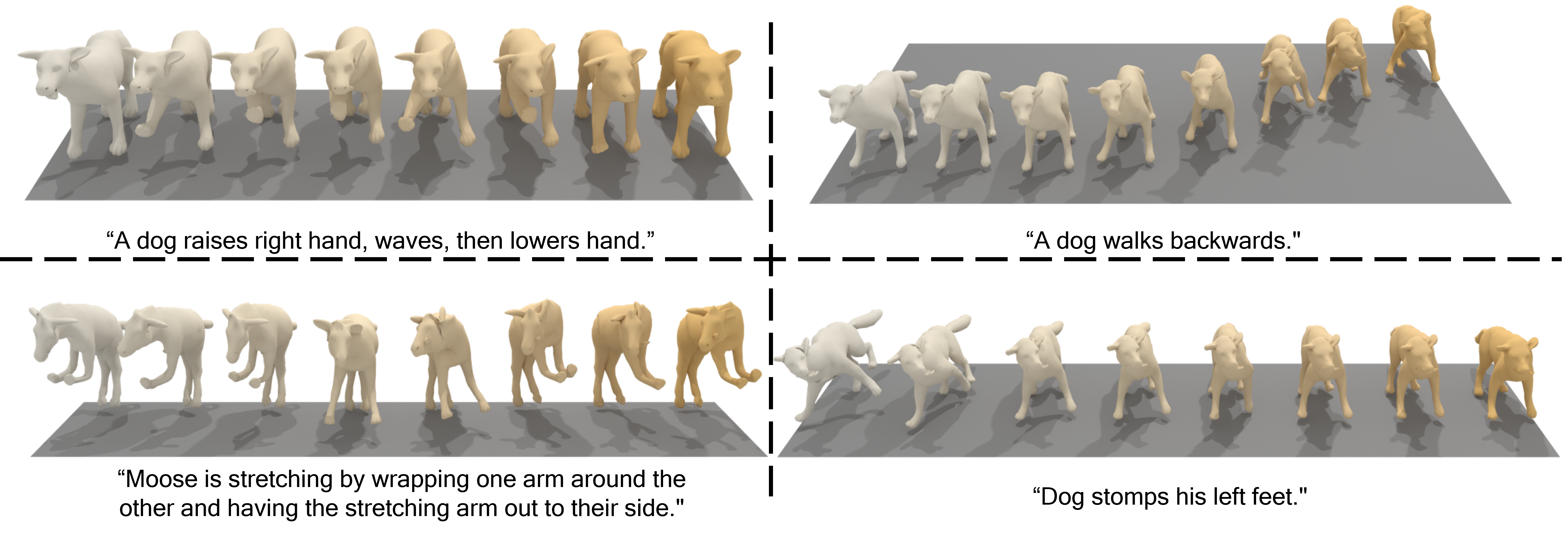}
\caption{
\textbf{More results of generated motions from our model.}
Our model demonstrates robust performance in generating both ID and OOD motions. 
Except for walking backward, all evaluated motions are OOD, underscoring the model's effectiveness in handling a variety of challenging scenarios.
}
\label{fig:supp_more_our_results}
\end{figure*}

In Figure~\ref{fig:supp_more_our_results}, we present additional motions generated by our OMGPT model. 
These results further validate our model's capability to generate both ID and OOD. For instance, walking backward is categorized as ID, while stomping with the left foot is considered OOD. 
A notable challenge is the generation of motions involving complex body interactions, such as stretching one arm with the assistance of the other. This aspect represents a critical area for future development, particularly in translating human motion interactions to animal models. Supplementary material, including a video that showcases these motions in a continuous format, is available. This video, named after the figures in this paper, provides a comprehensive view of the generated motions.

\section{Baseline Implementations}

For all baseline comparisons, we trained the models using our dataset, converting motions into a 36 by 6 dimensional format (details in Section~\ref{sec:ae}). These baseline models, originally designed for human motion generation, do not typically account for offsets, which are crucial in animal motion generation. Therefore, we incorporate offsets into the dynamic features as an additional input and output target. During inference, we directly use animal offsets for a fair comparison with our method. We adhere to the default settings provided in the baseline methodologies for both training and evaluation, ensuring consistency across all comparisons.

\section{More Baseline Results}

\begin{figure*}[t]
\centering
\includegraphics[width=0.92\linewidth]{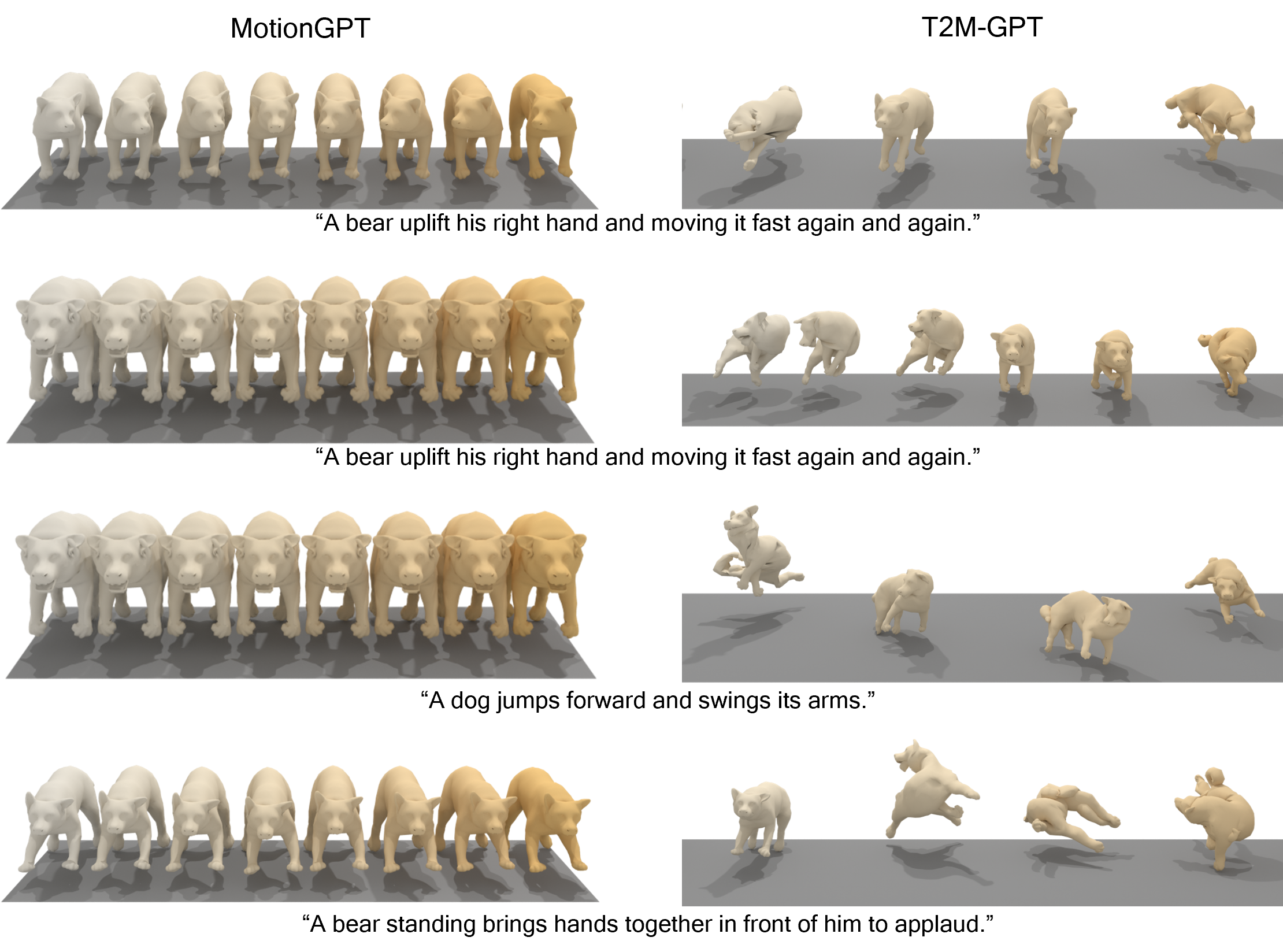}
\caption{
\textbf{Generated motions from T2M-GPT and MotionGPT.}
Figure illustrates motions generated by T2M-GPT~\cite{zhang2023generating} and MotionGPT~\cite{jiang2023motiongpt}, corresponding to comparisons in Figure~\ref{fig:baseline}. 
These results demonstrate comparatively lower quality, as evidenced by reduced metrics in R-Precision and MM-Dist.
}
\label{fig:supp_more_baselines}
\end{figure*}

In Figure~\ref{fig:supp_more_baselines}, we present results from T2M-GPT and MotionGPT. The analysis reveals that both models struggle with generating accurate motions: MotionGPT often produces motionless outputs in response to OOD inputs, whereas T2M-GPT tends to generate erratic and noisy motions under similar OOD conditions. 
This discrepancy highlights the challenge of aligning motion generation with the corresponding textual descriptions, especially when handling OOD instructions.

\section{Metric Computation Details}

We elaborate on several evaluation metrics, previously utilized in~\cite{guo2022t2m}. The metrics involve three types of features: ground-truth motion features ($f_{gt}$), generated motion features ($f_{pred}$), and text features ($f_{text}$). 
These features are extracted using the animal encoder, denoted as $E^a$, following the training of the network.

\paragraph{FID (Fréchet Inception Distance).}
This metric assesses the overall quality of generated motions. The FID is calculated using the equation:
\begin{equation}
\text{FID} = \lVert \mu_{gt} - \mu_{pred}\rVert^2 - \text{Tr}(\Sigma_{gt} + \Sigma_{pred} - 2(\Sigma_{gt}\Sigma_{pred})^{\frac{1}{2}})
\label{formula:fid}
\end{equation}
where $\mu_{gt}$ and $\mu_{pred}$ are mean of $f_{gt}$ and $f_{pred}$. $\Sigma$ is the covariance matrix and $\text{Tr}$ denotes the trace of a matrix. 
we calculate FID based on 1024 randomly generated motions.

\paragraph{MM-Dist.}
This metric calculates the feature-level distance between text embeddings and generated motion features. For N randomly generated samples, MM-Dist is the average Euclidean distance between each text feature and its corresponding generated motion feature, defined as:
\begin{equation}
\text{MM-Dist} = \frac{1}{N}\sum_{i=1}^{N}\lVert f_{pred,i} - f_{text,i}\rVert
\label{formula:mm-dis}
\end{equation}
where $f_{pred,i}$ and  $f_{text,i}$ are the features of the i-th text-motion pair. 
We set $N$ to 1024 in our experiments.

\paragraph{Diversity.} Diversity quantifies the variance among all motion sequences in the dataset. We calculate this by randomly selecting $S_{dis}$ pairs of motion features ($f_{pred,i}$ and $f_{pred,i}'$) and then computing:
\begin{equation}
\text{Diversity} = \frac{1}{S_{dis}}\sum_{i=1}^{S_{dis}}||f_{pred,i} - f_{pred,i}'||
\label{formula:diversity}
\end{equation}
$S_{dis}$ is set to 1024 for OOD and 64 for ID.

\paragraph{MModality.} this metric evaluates the diversity of human motions generated from the same text description. For each text description, we generate 100 motions and select two subsets containing 10 motions each. The features of the j-th pair for the i-th text description are denoted as ($f_{pred,i,j}$, $f_{pred,i,j}'$). MModality is then defined as:
\begin{equation}
\text{MModality} = \frac{1}{10N}\sum_{i=1}^{N}\sum_{j=1}^{10}\lVert f_{pred,i,j} - f_{pred,i,j}'\rVert
\label{formula:mmodality}
\end{equation}

\end{document}